\documentclass{article}

\usepackage[numbers]{natbib}
\usepackage[preprint]{neurips_2025_custom}

\usepackage{amsmath,amsfonts,bm}

\def\eqref#1{equation~(\ref{#1})}
\def\1{\bm{1}}

\DeclareMathAlphabet{\mathsfit}{\encodingdefault}{\sfdefault}{m}{sl}
\SetMathAlphabet{\mathsfit}{bold}{\encodingdefault}{\sfdefault}{bx}{n}

\usepackage[dvipsnames]{xcolor}         % colors
\definecolor{linkColor}{rgb}{0.18,0.39,0.62}
\usepackage[utf8]{inputenc} % allow utf-8 input
\usepackage[T1]{fontenc}    % use 8-bit T1 fonts
\usepackage[colorlinks=true,linkcolor=linkColor,citecolor=linkColor,filecolor=linkColor,urlcolor=linkColor]{hyperref}       % \usepackage{url}
\usepackage{cleveref}
\usepackage{multirow}
\usepackage{xspace}
\usepackage{booktabs}
\usepackage{graphicx}
\usepackage{array}
\usepackage{wrapfig}
\usepackage{bm}
\usepackage{algorithm}
\usepackage[font=small,labelfont=bf]{caption}

\usepackage{algpseudocode}
\usepackage{algpseudocode}
\usepackage{enumitem}
\usepackage{tcolorbox}
\usepackage{makecell}
\usepackage{diagbox}
\usepackage{arydshln}

\usepackage{amsmath}
\usepackage{amssymb}
\usepackage{mathtools}
\usepackage{amsthm}

\usepackage{multirow}
\usepackage{amsmath}
\usepackage{capt-of}
\usepackage{tabularx}
\usepackage{epsfig}
\usepackage{amssymb}
\usepackage{amsfonts}
\usepackage{booktabs}
\usepackage{scalerel}
\usepackage{listings}
\usepackage{varwidth}
\usepackage{stmaryrd}
\usepackage{bbm}
\usepackage{wrapfig}
\usepackage{pifont}

\definecolor{deepblue}{rgb}{0,0,0.5}
\definecolor{officeblue}{RGB}{0,102,204}
\definecolor{deepred}{rgb}{0.6,0,0}
\definecolor{deepgreen}{rgb}{0,0.5,0}
\definecolor{mybrickred}{RGB}{182,50,28}

\definecolor{fillcolor}{RGB}{216,217,252}

\usepackage{etoolbox}
\usepackage{framed}

\newif\ifxetexorluatex
\ifxetex
  \xetexorluatextrue
\else
  \ifluatex
    \xetexorluatextrue
  \else
    \xetexorluatexfalse
  \fi
\fi
\newcommand*\quotesize{60} % if quote size changes, need a way to make shifts relative
\newcommand*{\openquote}
   {\tikz[remember picture,overlay,xshift=-4ex,yshift=-2.5ex]
   \node (OQ) {\fontsize{\quotesize}{\quotesize}\selectfont``};\kern0pt}

\newcommand*{\closequote}[1]
  {\tikz[remember picture,overlay,xshift=4ex,yshift={#1}]
   \node (CQ) {\fontsize{\quotesize}{\quotesize}\selectfont''};}

\colorlet{shadecolor}{white}

\newcommand*\shadedauthorformat{\emph} % define format for the author argument

\newcommand*\authoralign[1]{%
  \if#1l
    \def\authorfill{}\def\quotefill{\hfill}
  \else
    \if#1r
      \def\authorfill{\hfill}\def\quotefill{}
    \else
      \if#1c
        \gdef\authorfill{\hfill}\def\quotefill{\hfill}
      \else\typeout{Invalid option}
      \fi
    \fi
  \fi}
{\authoralign{#1}
\ifblank{#2}
   {\def\shadequoteauthor{}\def\yshift{-2ex}\def\quotefill{\hfill}}
   {\def\shadequoteauthor{\par\authorfill\shadedauthorformat{#2}}\def\yshift{2ex}}
\begin{snugshade}\begin{quote}\openquote}
{\shadequoteauthor\quotefill\closequote{\yshift}\end{quote}\end{snugshade}}

\definecolor{DarkBlue}{RGB}{0, 51, 153}
\newcommand{\oursattn}{CLSA}
\newcommand{\ours}{YOCO (CLSA)}
\newcommand{\oursdense}{YOCO (Dense)}

\title{You Only Index Once: \\Cross-Layer Sparse Attention with Shared Routing}

\author{
Yutao Sun\thanks{~Equal contribution.}~~$^{12}$~~~~Yanqi Zhang\footnotemark[1]~~$^{1}$~~~~Li Dong\footnotemark[1]~~$^{1}$ \\
\bf Jianyong Wang$^{2}$~~~~Furu Wei$^{1}$ \\
$^1$ Microsoft Research ~~~~
$^2$ Tsinghua University \\
{\href{https://aka.ms/GeneralAI}{https://aka.ms/GeneralAI}}
}

\begin{document}

\maketitle

\begin{abstract}
\noindent Long-context inference in modern LLMs is increasingly constrained by decoding efficiency, especially in reasoning-heavy settings where models generate long intermediate chains of thought. Existing sparse attention methods often face a practical efficiency-quality trade-off. Structured block sparse methods typically provide stronger acceleration but incur noticeable quality loss, while token sparse methods are usually more accurate yet deliver limited end-to-end speedup because top-$k$ routing over the full cache remains expensive. In this work, we propose cross-layer sparse attention (CLSA), which is built on top of KV-sharing architectures such as YOCO. The core idea is to share not only the KV cache across cross-decoder layers, but also the routing index. A single indexer computes token-level top-$k$ selection once and reuses the resulting index across layers, thereby preserving the fine-grained selectivity of token sparse attention while amortizing the routing overhead. The resulting architecture improves all major inference bottlenecks jointly, including pre-filling, KV-cache storage, and long-context decoding. Experiments across short-context and long-context benchmarks show that \oursattn{} is both accurate and efficient, achieving up to $7.6\times$ decoding speedup and $17.1\times$ overall throughput improvement at 128K context. These results suggest a more complete architectural solution for long-context LLMs that jointly advances model quality and inference efficiency.
\end{abstract}

\section{Introduction}

Long-context inference has become a common operating regime for modern LLMs, especially in reasoning-heavy settings such as chain-of-thought generation and test-time scaling~\citep{o1,deepseekr1}. In these scenarios, models often need to decode long intermediate reasoning traces while repeatedly attending to a large context, making inference increasingly decoding-bound as the sequence grows. At the same time, pre-filling becomes more expensive and the KV cache grows with context length. Designing an architecture that remains efficient under long contexts is therefore now a core requirement rather than a niche optimization.

Sparse attention is a natural direction for reducing the cost of long-context inference, but existing methods often face a difficult efficiency-quality trade-off. In practice, block-sparse attention~\citep{nsa,moba,quest,seerattention} usually delivers larger wall-clock speedups because its structured sparsity maps better to GPU execution, but it also tends to introduce a coarser approximation and more noticeable quality loss. Token-sparse attention~\citep{deepseek3.2} is often more accurate because it can preserve finer-grained salient tokens, yet its end-to-end acceleration is usually limited. A key reason is that token sparse methods still require a routing stage based on top-$k$ selection over the full cache, and this step is irregular and expensive on modern GPUs, especially when it is recomputed independently across many layers during decoding.

In this work, we propose cross-layer sparse attention (CLSA), which is built on top of the KV sharing design of cross-attention architectures such as YOCO~\citep{yoco}. The central idea is to extend sharing from memory to routing. When multiple cross-decoder layers read from the same KV cache, they also share the same routing index. Concretely, a single indexer computes token-level top-$k$ routing once and the resulting index is reused across layers. In this way, the model preserves the main advantage of token-sparse attention, namely selecting a compact active subset of informative tokens without sacrificing quality, while substantially reducing the practical cost of sparse decoding by amortizing the routing overhead.

The resulting architecture provides a unified treatment of the major inference bottlenecks in long-context LLMs. Through KV sharing, it retains YOCO's advantages in pre-filling and KV-cache storage, while shared-index sparse retrieval improves decoding efficiency by avoiding repeated dense global attention and repeatedly recomputed routing. Consequently, the overall system approaches a favorable efficiency frontier across pre-filling, KV-cache footprint, and long-context decoding, rather than improving only one aspect at the expense of the others.

Our experiments show that \oursattn{} preserves model quality across both short and long context benchmarks spanning multiple domains, while maintaining nearly lossless behavior relative to dense baselines. Attention-pattern analysis further shows that sharing one routing index across layers has only a minor effect on the resulting attention behavior, supporting the central assumption behind our design. At the same time, \oursattn{} delivers substantial acceleration. At 128K context, it improves decoding throughput by up to $7.6\times$ over the Transformer baseline and improves overall end-to-end throughput by up to $17.1\times$. Taken together, these results suggest that \oursattn{} provides a potentially promising LLM architecture that better reconciles model quality and inference efficiency.

\section{Method}

\begin{figure}[!t]
\centering
\includegraphics[width=0.9\linewidth]{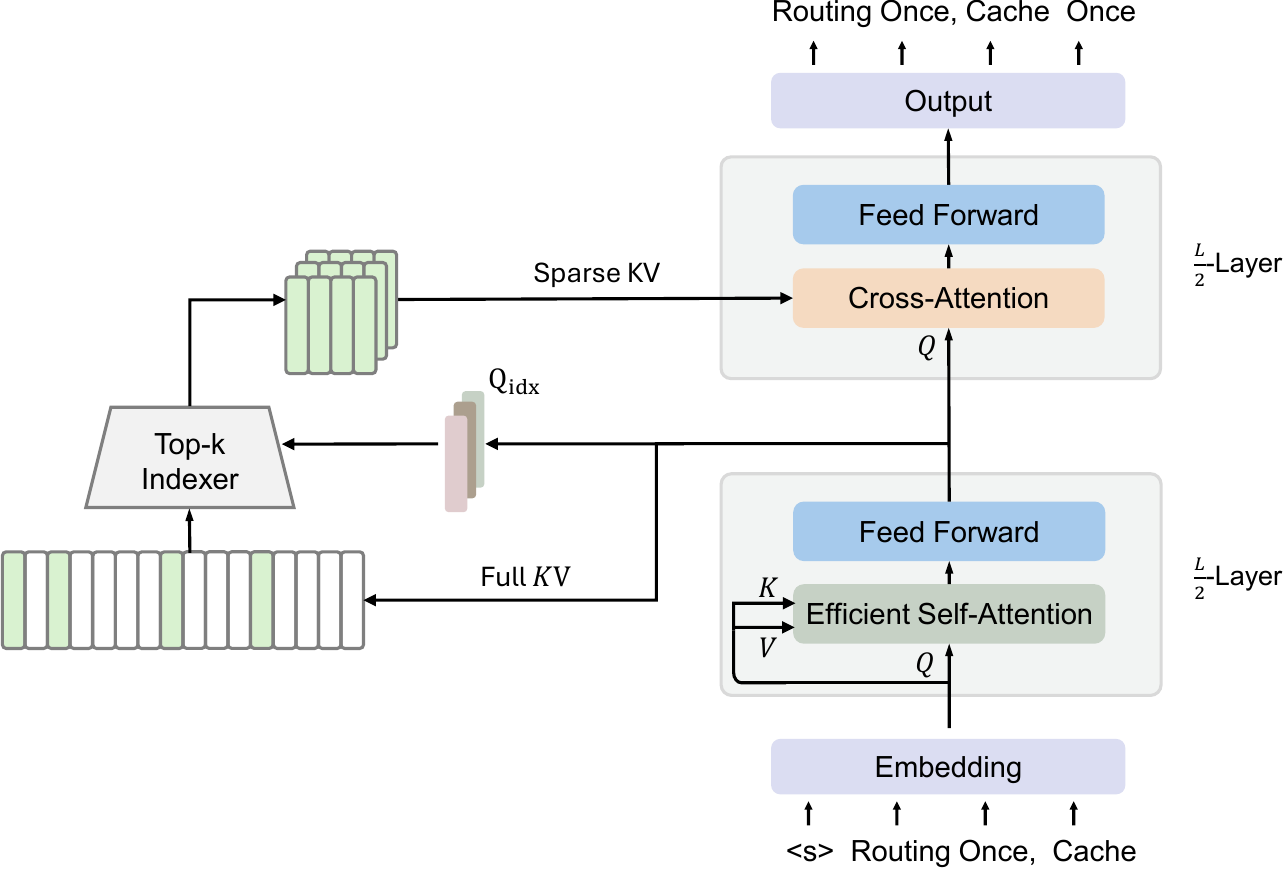}
\caption{Overview of cross-layer sparse attention. The self-decoder first produces a shared KV cache, which is computed only once and then reused by all subsequent cross-decoder layers. During this stage, a shared query-aware indexer jointly generates the routing queries and keys and computes a token-level sparse top-$k$ index for each query token. This sparse index is also produced only once and is shared across the following cross-decoder layers, allowing them to reuse the same selected KV positions instead of recomputing layer-specific sparse routing.}
\label{fig:method}
\end{figure}

\Cref{fig:method} gives an overview of our method. We build on the cross-attention architecture of YOCO~\citep{yoco}, which naturally decomposes the model into a self-decoder and a cross-decoder. The self-decoder first encodes the input sequence into shared hidden states and constructs a single KV cache. On top of these shared states, we add a lightweight query-aware indexer that computes a token-level top-$k$ routing index once. The cross-decoder layers then reuse both the shared KV cache and the shared routing index, so each layer keeps its own query states and FFN transformation while attending only to the selected KV positions.

This overview highlights the main design principle of cross-layer sparse attention: when several decoder layers read from the same memory, the expensive routing decision should also be tied to that memory and shared across layers. As a result, CLSA preserves the fine-grained selectivity of token-sparse attention, but avoids recomputing layer-specific top-$k$ indices during decoding. The following subsections detail the sparse attention formulation, the multi-layer distillation objective used to train the shared indexer, and the resulting inference advantages.

\paragraph{Efficient Self-decoder.}
The self-decoder is kept unchanged from YOCO. It performs an efficient attention mechanism and constructs the shared KV cache exactly once. This stage encodes the history into reusable memory while avoiding dense full-context attention throughout the whole stack. As a result, the model preserves YOCO's efficiency advantages in both pre-filling and KV-cache storage.

\paragraph{Sparse Cross-decoder.}
Each cross-decoder layer consists of a cross-attention module followed by an FFN block. Its role is to retrieve relevant information from the shared KV cache and refine the representation through the feed-forward transformation. In dense YOCO, the cross-attention module performs dense full attention over the shared cache. Our modification is to replace this dense retrieval with CLSA driven by the shared indexer, while keeping the FFN and the rest of the cross-decoder unchanged.

\subsection{Cross-Layer Sparse Attention}
As shown in \Cref{fig:method}, we consider a stack of decoder layers that all query the same external memory through a cross-attention module. Let $H \in \mathbb{R}^{n \times d}$ denote the shared hidden states written into the memory, let $K,V \in \mathbb{R}^{n \times d}$ denote the corresponding shared KV cache, and let $Q^{(l)}$ be the query states of decoder layer $l$. A dense cross-attention module computes
\begin{equation}
O^{(l)} = \mathrm{Attn}(Q^{(l)}, K, V),
\end{equation}
which means every decoder layer scans the full cache. Our goal is to keep the shared-memory structure, but replace dense retrieval with a routed sparse variant. To do so, we introduce an additional single-head indexing branch on top of the shared hidden states. Specifically, the index queries and keys are computed as:
\begin{equation}
Q_{\mathrm{idx}} = H W^{Q}_{\mathrm{idx}}, \ \ 
K_{\mathrm{idx}} = H W^{K}_{\mathrm{idx}},
\end{equation}
where $Q_{\mathrm{idx}}, K_{\mathrm{idx}} \in \mathbb{R}^{n \times d_{\mathrm{idx}}}$. Unlike the multi-head attention projections used by the main attention module, this indexing branch uses only one head. We then form indexing scores and routing indices
\begin{equation}
I = Q_{\mathrm{idx}} K_{\mathrm{idx}}^\top, \ \ 
S_t = \mathrm{TopK}(I_t, k)
\end{equation}
and restrict attention to the selected tokens:
\begin{equation}
O_t^{(l)} = \mathrm{Attn}(Q_t^{(l)}, K_{S_t}, V_{S_t})
\end{equation}
Here the index query and index key are both derived from the same shared hidden states and have shape $[n, d_{\mathrm{idx}}]$, while the layer-wise query $Q^{(l)}$ still varies across decoder layers. The activated set size $k$ is much smaller than $n$, so each layer only attends to a compact subset of the global memory.

The central idea is to bind the routing index to the shared KV cache, rather than treating them as separate objects. Once multiple decoder layers attend to the same KV cache, they should also reuse the same routing index, so inference only needs to compute top-$k$ once.

\subsection{Multi-Layer Distillation}
Sharing one routing index across multiple decoder layers is efficient, but the selected tokens must be useful to several layers simultaneously rather than matching the preference of one layer in isolation. Follow prior sparse attention methods with distillation~\citep{seerattention,indexcache}, we use a multi-layer distillation objective to train the shared indexer.

For an input sequence, the dense cross-attention modules provide attention distributions from all decoder layers and all heads. We first aggregate them into a common target
\begin{equation}
\bar{A}
=
\frac{1}{LH}\sum_{l=1}^{L}\sum_{h=1}^{H}
\mathrm{softmax}\left(Q^{(l,h)} {K^{(h)}}^\top\right),
\end{equation}
and then match that target with the indexer distribution using
\begin{equation}
\mathcal{L}_{\mathrm{KD}}
=
\frac{1}{n}\sum_{t=1}^{n}
\mathrm{KL}\left(
\mathrm{sg}\left[\bar{A}_t\right]
\;\middle\|\;
\mathrm{softmax}(I_t)
\right),
\end{equation}
where $L$ is the number of decoder layers, $H$ is the number of attention heads in each layer, and $\mathrm{sg}[\cdot]$ denotes stop-gradient. Intuitively, the shared indexer learns to preserve the consensus salient tokens that remain important across the full decoder stack.

\paragraph{Stage 1: indexer warmup with frozen backbone.}
We first warm up the shared indexer using only the distillation objective while freezing the backbone parameters. In this stage, the indexer learn a stable routing pattern before it is coupled with language modeling.
\begin{equation}
\mathcal{L}_{\mathrm{stage1}} = \mathcal{L}_{\mathrm{KD}}.
\end{equation}

\paragraph{Stage 2: joint sparse adaptation.}
After the indexer warmup, we optimize the model with both language modeling and distillation losses. The main purpose of this stage is to let the backbone adapt to the sparse attention distribution induced by the shared indexer:
\begin{equation}
\mathcal{L}_{\mathrm{stage2}}
=
\mathcal{L}_{\mathrm{LM}}
+ \lambda \mathcal{L}_{\mathrm{KD}}.
\end{equation}
where \(\lambda\) is a fixed weighting coefficient for the KD loss, and we set \(\lambda=0.1\) in all experiments.

\subsection{Inference Advantages}

Decoding efficiency, pre-filling efficiency, and KV-cache footprint cover most of the practical bottlenecks in LLM inference. Our design improves these three aspects in a complementary way.

\begin{table}[t]
\centering
\caption{Comparison of inference complexity. $N$, $L$, and $D$ denote sequence length, number of layers, and hidden dimension. $W_1$ denotes the local window size, and $W_2$ denotes the number of selected tokens in sparse attention. $\gamma$ denotes the fraction of global-attention layers in a hybrid model, and $\eta$ denotes the per query-key routing cost of the indexer.}
\label{tbl:complexity:combined}
\vspace{0.5em}
\resizebox{\linewidth}{!}{
\begin{tabular}{llll}
\toprule
\textbf{Model} & \textbf{KV Cache Memory} & \textbf{Prefilling Time} & \textbf{Decoding Time} \\
\midrule
Transformer & $\mathcal{O}(LND)$ & $\mathcal{O}(LN^2D)$ & $\mathcal{O}(LND)$ \\
Hybrid TRM & $\mathcal{O}(L(\gamma N + (1-\gamma)W_1)D)$ & $\mathcal{O}(L(\gamma N^2 + (1-\gamma)W_1N)D)$ & $\mathcal{O}(L(\gamma N + (1-\gamma)W_1)D)$ \\
\oursdense{} & $\mathcal{O}((N+W_1L)D)$ & $\mathcal{O}(\frac{L}{2}W_1ND)$ & $\mathcal{O}(\frac{L}{2}(N+W_1)D)$ \\
DSA & $\mathcal{O}(LND)$ & $\mathcal{O}(LW_2ND+\eta LN^2)$ & $\mathcal{O}(LW_2D+\eta LN)$ \\
\ours{} & $\mathcal{O}((N+W_1L)D)$ & $\mathcal{O}(\frac{L}{2}W_1ND)$ & $\mathcal{O}(\frac{L}{2}(W_1+W_2)D+\eta N)$ \\
\bottomrule
\\
\end{tabular}
}
\end{table}

\paragraph{Decoding efficiency from shared top-$k$ routing.}
In a standard sparse design, each decoder layer still needs to run its own routing procedure to identify the top-$k$ tokens, and that routing must still be computed over the full-length cache. More importantly, the top-$k$ operator is not well matched to modern GPU execution. Unlike dense matrix multiplications, it cannot effectively leverage Tensor Core acceleration, so its wall-clock cost can account for a large fraction of decoding time. Our shared-indexer design removes this redundancy by computing the top-$k$ result once and reusing it across the decoder stack. As a result, sparse retrieval becomes practically useful for decoding. While the model still attends to only a small active set, but inference no longer wastes time recomputing the same expensive top-$k$ step in every layer.

\paragraph{Pre-filling and KV-cache efficiency inherited from YOCO.}
YOCO~\citep{yoco} already makes pre-filling efficient by avoiding dense full-context attention, and it reduces KV-cache memory by letting the cross-decoder layers reuse a single shared cache. Our method preserves these benefits completely, because it only changes how the shared cache is read by the cross-attention modules in the cross-decoder. The two components are complementary: YOCO addresses pre-filling and KV-cache storage, while cross-sparse attention addresses decoding. Together they cover nearly all major bottlenecks in LLM inference, yielding a unified design that is efficient in pre-filling, memory footprint, and long-context generation.

\paragraph{Complexity comparison with other architectures.}
\Cref{tbl:complexity:combined} highlights why improving only one component is insufficient for long-context inference. Pure dynamic sparse attention~\cite{deepseek3.2} reduces the number of tokens read by each attention layer, but it still maintains a layer-wise KV cache of size $\mathcal{O}(LND)$ and therefore does not address the memory bottleneck. Its decoding cost can also be dominated by the indexer term $\eta LN$, because token selection must be performed over the full cache and repeated across layers. Hybrid architectures~\cite{qwen3,kimi-linear,hysparse} reduce part of the attention cost by mixing efficient-attention layers with global retrieval layers, but their gains are constrained by the hybrid ratio $\gamma$. In contrast, \oursattn{} combines KV sharing with a shared routing index, reducing cache storage through YOCO-style memory sharing while paying the expensive indexer cost only once across the cross-decoder stack.

\begin{table}[!t]
    \centering
    \small
    \caption{Main downstream benchmark results for the 4B Transformer baseline, dense YOCO, and YOCO with cross-sparse attention. \ours{} maintains the overall capability profile of the dense baselines while improving ARC-Challenge (ARC-C), GSM8K, and DROP, and matching the best HumanEval score.}
    \label{tab:main_results}
    \vspace{0.5em}
    \resizebox{\linewidth}{!}{
    \begin{tabular}{@{}lcccccccc@{}}
    \toprule
    \textbf{Model} & \textbf{ARC-C} & \textbf{BBH} & \textbf{GSM8K} & \textbf{HellaSwag} & \textbf{HumanEval} & \textbf{MMLU} & \textbf{DROP} & \textbf{WinoGrande} \\
    \midrule
    Transformer & 0.453 & \textbf{0.420} & 0.434 & 0.667 & 0.384 & \textbf{0.527} & 0.366 & \textbf{0.638} \\
    \oursdense{}        & 0.461 & 0.411 & 0.430 & \textbf{0.676} & \textbf{0.396} & 0.519 & 0.387 & 0.630 \\
    \ours{}     & \textbf{0.465} & 0.418 & \textbf{0.470} & 0.674 & \textbf{0.396} & 0.513 & \textbf{0.391} & 0.616 \\
    \bottomrule
    \\
    \end{tabular}
    }
    \end{table}

\section{Experiments}

\subsection{Setup}

\paragraph{Model configuration}
We compare a Transformer baseline, a dense YOCO model, and YOCO with cross-sparse attention, all at the 4B scale.
The baseline uses RoPE with \texttt{base}${=}500{,}000$, and both YOCO variants use RNoPE~\citep{rnope}, where RoPE~\citep{rope} is activated in SWA with \texttt{base}${=}10{,}000$ and NoPE in global attention. Both YOCO variants use sliding window attention in the self-decoder with window size $512$. QK normalization is enabled in the Transformer and the self-decoder of YOCO. The maximal activated tokens in \oursattn{} is $2048$.
Across models we keep width and depth aligned with hidden size $2560$, $7680$ FFN width, $32$ layers, $20$ heads and $4$ KV heads, no weight tying. For the YOCO variants, the 32 layers are split into 16 self-decoder layers and 16 cross-decoder layers. The full field-by-field layout is in \Cref{app:model-config}.

\paragraph{Training hyper-parameters}
Dense pretraining runs in two stages.
In dense stage~1 we use batches of 8M tokens per step, maximum sequence length $8192$, peak learning rate $3{\times}10^{-4}$ with minimum $3{\times}10^{-5}$, $2000$ warmup iterations, and $125{,}000$ optimizer updates.
Dense stage~2 increases the context cap to $32{,}768$ tokens, fixes the learning rate at $3{\times}10^{-5}$, and runs for $10{,}000$ optimizer steps.
Sparse adaptation also uses two stages on $32{,}768$-token sequences: sparse stage~1 likewise uses 8\,M tokens per step, learning rate $3{\times}10^{-4}$ with the same minimum $3{\times}10^{-4}$, for $2500$ steps, and sparse stage~2 continues for another $2500$ steps at $3{\times}10^{-5}$.
Further settings and stage-wise details are given in \Cref{app:training-hparams}.

\paragraph{Evaluation benchmarks}
We evaluate models with BBH~\citep{bbh} and MMLU~\citep{mmlu} for heterogeneous knowledge and reasoning, DROP~\citep{drop} and ARC-Challenge~\citep{arc} for reading-style reasoning, HellaSwag~\citep{hella_swag} and WinoGrande~\citep{winogrande} for commonsense multiple-choice completion, GSM8K~\citep{gsm8k} for grade-school math word problems, HumanEval~\citep{humaneval} for Python function synthesis, and RULER~\citep{ruler} for long-context synthetic retrieval.

\subsection{General Benchmark}
\Cref{tab:main_results} summarizes the main downstream results for the Transformer baseline, \oursdense{}, and \ours{}. Overall, \oursattn{} preserves the broad capability profile of the dense models while improving performance on several tasks that require selective evidence aggregation. In particular, \ours{} obtains the best scores on ARC-Challenge, GSM8K, and DROP, and matches the best HumanEval result. On BBH, MMLU, HellaSwag, and WinoGrande, its performance remains close to the dense baselines, indicating that sparsifying the global attention path does not introduce a systematic degradation of general reasoning or knowledge. More broadly, hybrid architectures may also yield quality gains by combining complementary modeling capabilities and by leveraging multiple positional-encoding configurations.

\subsection{Long Context}

\begin{table*}[t]
\centering
\small
\setlength{\tabcolsep}{2pt}
\caption{RULER results at 16K and 32K context lengths. TRM denotes the standard Transformer. \oursattn{} maintains strong single-needle retrieval performance and achieves the best average score at 32K, with gains mainly from the harder multi-needle settings.}
\label{tab:ruler_results}
\vspace{0.5em}
\begin{tabular}{ll|ccccccccccccc}
\toprule
\textbf{Ctx} & \textbf{Model}
& \multicolumn{3}{c}{\textbf{Single Needle}}
& \multicolumn{5}{c}{\textbf{Multi Needle}}
& \multicolumn{4}{c}{\textbf{RULER Tasks}}
& \textbf{Avg} \\
\cmidrule(lr){3-5}
\cmidrule(lr){6-10}
\cmidrule(lr){11-14}

&
& S1 & S2 & S3
& MK1 & MK2 & MK3 & MQ & MV
& QH & QS & CWE & FWE
& \\
\midrule

\multirow{3}{*}{\textsc{16K}}
& TRM
& 100.0 & 99.8 & 98.4
& 88.2 & 71.4 & 14.4 & 85.7 & 85.6
& 28.8 & 33.2 & 15.6 & 52.1
& \textbf{64.4} \\

& \oursdense{}
& 100.0 & 99.8 & 96.4
& 69.4 & 91.6 & 61.2 & 45.8 & 49.3
& 30.8 & 31.4 & 9.4 & 67.0
& 62.7 \\

& \ours{}
& 100.0 & 100.0 & 98.4
& 70.4 & 92.4 & 58.4 & 53.0 & 47.2
& 31.2 & 32.7 & 9.8 & 61.6
& 62.9 \\

\midrule

\multirow{3}{*}{\textsc{32K}}
& TRM
& 100.0 & 98.8 & 83.4
& 57.0 & 38.8 & 0.8 & 45.6 & 42.6
& 21.2 & 20.2 & 1.8 & 43.8
& 46.2 \\

& \oursdense{}
& 100.0 & 90.2 & 74.8
& 53.2 & 84.0 & 43.6 & 27.0 & 29.0
& 30.6 & 30.6 & 4.6 & 60.3
& 52.3 \\

& \ours{}
& 100.0 & 93.6 & 83.2
& 58.4 & 88.8 & 38.0 & 31.6 & 29.8
& 29.2 & 29.2 & 5.1 & 50.2
& \textbf{53.1} \\

\bottomrule
\\
\end{tabular}
\end{table*}

\Cref{tab:ruler_results} reports long-context results on RULER at 16K and 32K tokens. At 16K, the Transformer baseline achieves the best overall average, while \oursattn{} remains competitive with dense attention and preserves near-perfect performance on the single-needle retrieval tasks. At 32K, where long-range interference becomes more pronounced, \oursattn{} achieves the best average score among all models. The improvement is mainly driven by stronger robustness on the more difficult multi-needle settings, especially MK1 and MK2, while maintaining comparable performance on the single-needle tasks.

\begin{figure}[t]
\centering
\includegraphics[width=\linewidth]{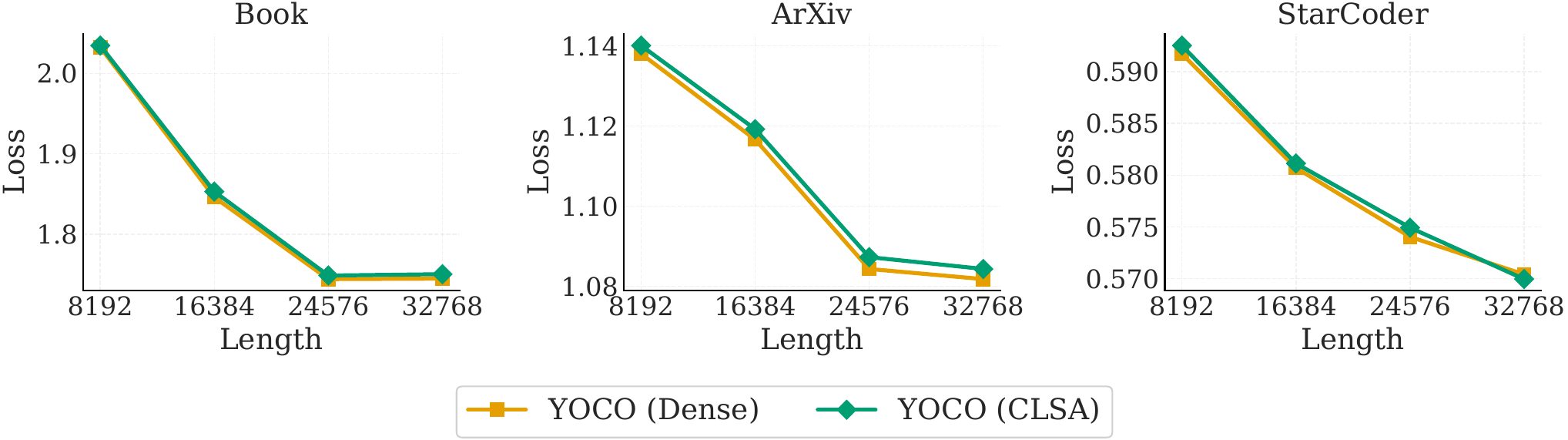}
\caption{Long-context validation loss for dense and cross-sparse attention on Books, ArXiv, and StarCoder. The two curves track each other closely from 8K to 32K tokens.}
\label{fig:long_seq}
\end{figure}

As shown in \Cref{fig:long_seq}, beyond these synthetic retrieval benchmarks, we also measure language modeling quality on long validation slices from Books, ArXiv, and StarCoder at context lengths from 8K to 32K tokens. \oursdense{} and \ours{} exhibit nearly overlapping cross-entropy loss across domains and lengths, indicating that \oursattn{} is effectively lossless for long-context modeling. As context grows from 8K to 32K, \oursattn{} follows the same loss trend as dense attention across all three domains. This indicates that CLSA preserves the same context-scaling behavior.

\subsection{Inference Efficiency}

\begin{figure}[!t]
\centering
\includegraphics[width=0.8\linewidth]{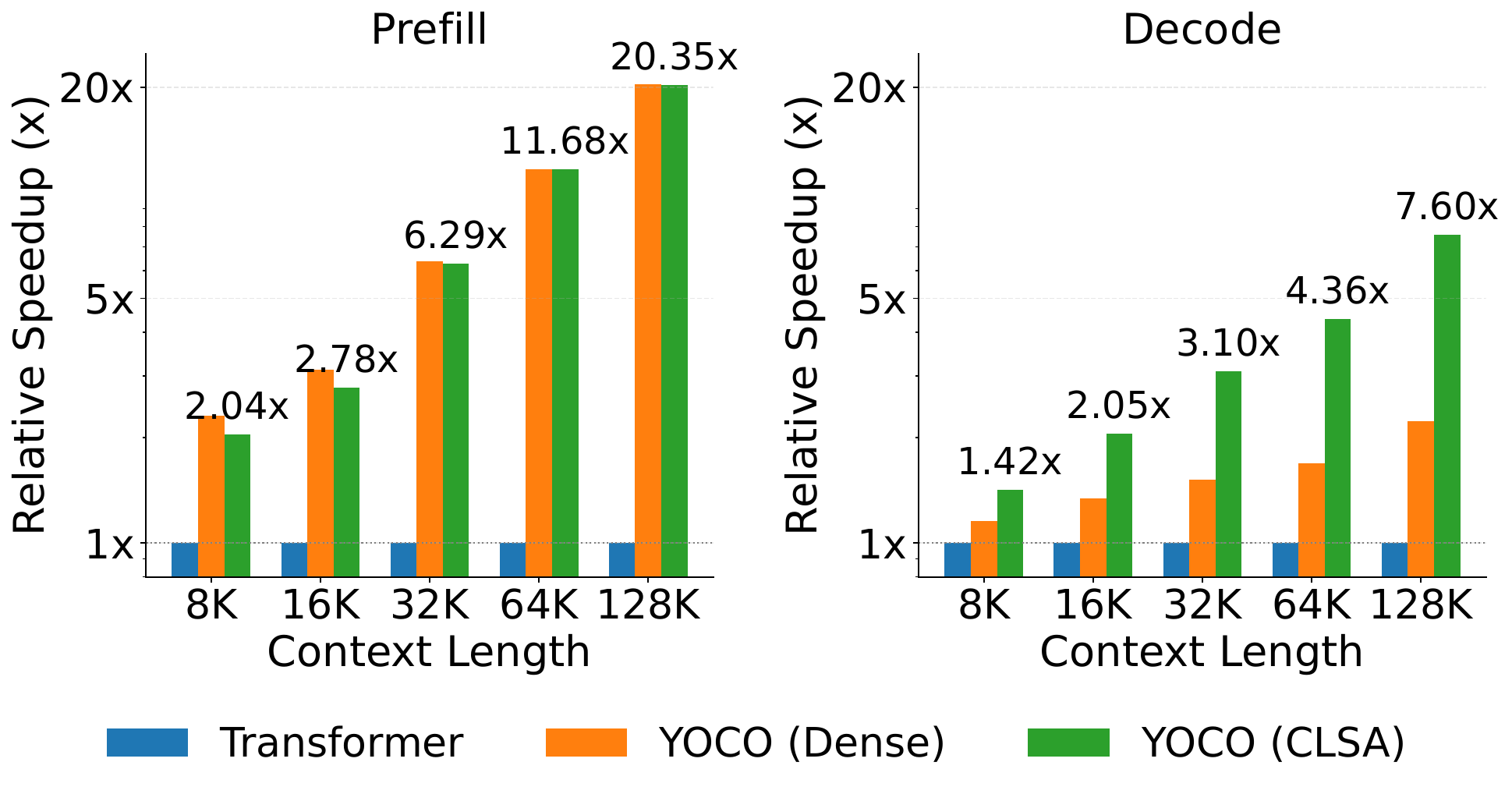}
\caption{Inference throughput relative to the Transformer for prefill and decode across different context lengths. Both YOCO variants substantially accelerate prefill, while \oursattn{} provides the largest decoding gains and widens its advantage as the context grows.}
\label{fig:inference}
\end{figure}

We integrate \oursattn{} into the open-source vLLM~\citep{vllm} inference stack by merging our implementation into its codebase, and we report end-to-end serving measurements from the resulting build. All numbers in this subsection are obtained on NVIDIA B200 GPUs.

\Cref{fig:inference} compares prefill and decode throughput across context lengths. The raw throughput values for the plotted prefill and decode panels, along with the overall end-to-end generation measurements, are listed in \Cref{app:exp-details-inference}. During prefill, both YOCO variants are substantially faster than the Transformer because the decoder architecture avoids quadratic full-context attention, and \ours{} remains close to \oursdense{}. The main benefit of cross-sparse attention appears during decoding, where \oursattn{} is consistently faster than both baselines, and the margin grows with context length. At 128K context, \oursattn{} achieves roughly $7.6\times$ Transformer decode throughput and about $17.1\times$ overall throughput, showing that sparsifying the global path translates directly into practical generation speed.

\begin{figure}[!t]
\centering
\begin{minipage}[t]{0.46\textwidth}
    \centering
    \includegraphics[width=0.8\linewidth]{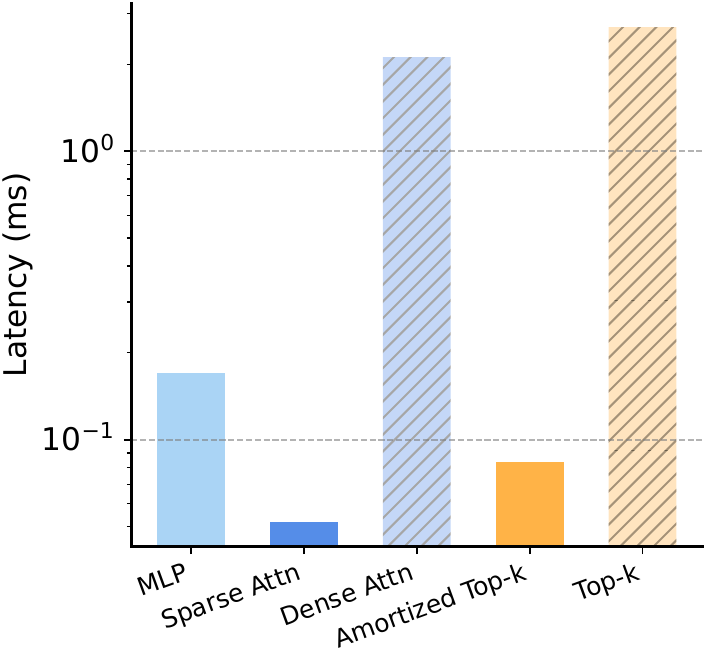}
    \caption{128K latency analysis for different components. After amortizing routing, the amortized top-$k$ becomes efficient. Without amortization, the unamortized top-$k$ stage can be comparable to or even larger than dense attention.}
    \label{fig:breakdown_128k}
\end{minipage}
\hfill
\begin{minipage}[t]{0.46\textwidth}
    \centering
    \includegraphics[width=0.8\linewidth]{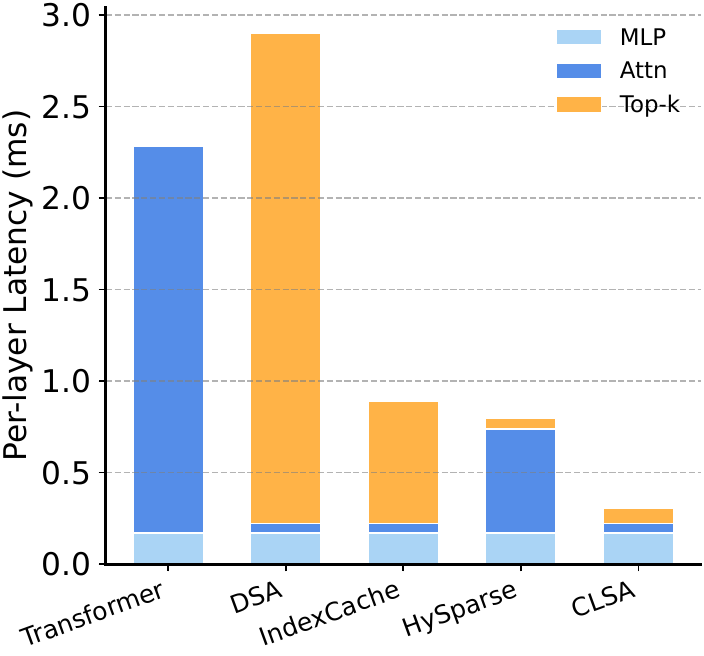}
    \caption{Per-layer latency comparison across representative sparse attention methods and dense baselines at 128K context. CLSA achieves the lowest latency by amortizing routing across cross-decoder layers.}
    \label{fig:breakdown_models_128k}
\end{minipage}
\end{figure}

\Cref{fig:breakdown_128k} demonstrates that top-$k$ routing is irregular and poorly matched to the wide, data-parallel execution that dense matrix multiplies exploit, so a standalone top-$k$ pass at 128K context can take time comparable to substantial dense attention work despite involving far fewer arithmetic operations. Sparse attention therefore only becomes practically faster when routing is amortized as in \oursattn{}, where the same routing decision is reused across multiple layers so the one-off top-$k$ cost is shared over the attention computations it replaces.

\Cref{fig:breakdown_models_128k} further compares per-layer decode latency against representative sparse attention methods, including DSA~\cite{deepseek3.2}, which performs independent token-level top-$k$ every layer, IndexCache~\cite{indexcache}, which reuses the routing index across four layers, and HySparse~\cite{hysparse}, which mixes block-sparse layers with a fraction of dense layers. At 128K context, DSA is slower than the dense Transformer because the unamortized token top-$k$ dominates per-layer cost. IndexCache and HySparse reduce this overhead but still pay substantial attention or routing costs. \oursattn{} achieves the lowest per-layer latency by combining lightweight sparse attention with fully amortized routing across all cross-decoder layers.

\begin{figure}[!t]
\centering
\includegraphics[width=\linewidth]{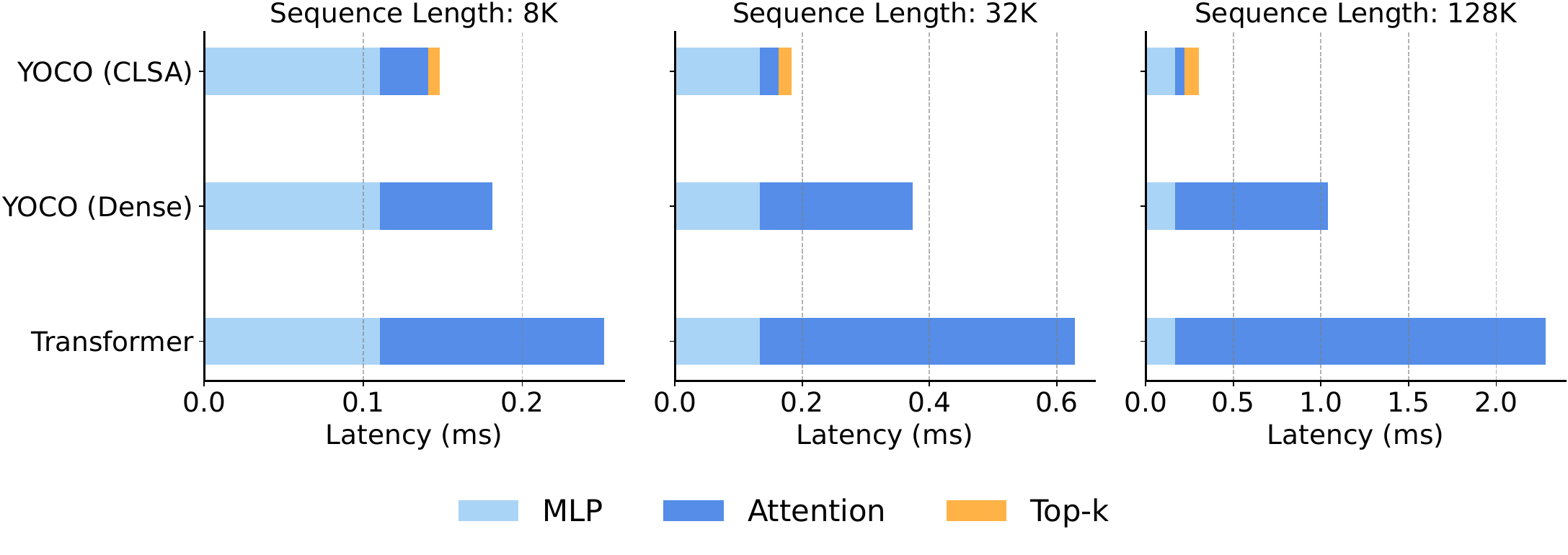}
\caption{Per-layer latency breakdown at 8K, 32K, and 128K context. For YOCO (Dense), the attention cost is averaged over SWA and dense attention layers. For \ours{}, the attention cost is averaged over SWA and CLSA layers, and the top-$k$ cost is amortized across cross-decoder layers. At 128K context, the amortized top-$k$ stage takes about $0.08$ ms per layer.}
\label{fig:breakdown}
\end{figure}

\Cref{fig:breakdown} decomposes per-layer latency, and the corresponding numeric values are listed in \Cref{app:exp-details-breakdown}. For YOCO (Dense), the reported attention term averages the cost of SWA in the 16 self-decoder layers and dense full attention in the 16 cross-decoder layers. For \ours{}, it averages the same 16 SWA layers and the 16 CLSA layers. The top-$k$ routing is executed once for the shared KV cache and its output is reused by the 16 cross-decoder layers. To make the stacked bars comparable across Transformer, \oursdense{}, and \ours{}, \Cref{fig:breakdown} reports all components after normalization by the full 32-layer model depth. Therefore, the plotted top-$k$ term is the one-off routing cost divided by 32, while the architectural sharing occurs across the 16 cross-decoder layers. In a dense Transformer stack, attention grows into the dominant term as sequences lengthen, whereas in \oursattn{} the sparse attention kernel itself stays comparatively light. Yet lower theoretical FLOPs do not automatically translate into wall-clock speedups on modern GPUs.

\subsection{Attention Sparsity Analysis}

\begin{table}[!t]
\centering
\small
\caption{Attention coverage and cross-entropy loss under sparse selection across selected-token budgets. Larger budgets recover more dense attention mass. Importantly, sparse selection introduces negligible cross-entropy loss degradation. 2048 selected tokens provide a favorable trade-off across domains.}
\label{tab:attention_analysis}
\vspace{0.5em}
\setlength{\tabcolsep}{4pt}
\begin{tabular}{l cccc cccc :cc}
\toprule
& \multicolumn{4}{c}{\textbf{Attn.\ Coverage (\%\,$\uparrow$)}} & \multicolumn{6}{c}{\textbf{Cross-Entropy Loss ($\downarrow$)}} \\
\cmidrule(lr){2-5} \cmidrule(lr){6-11}
\textbf{Domain} & \textbf{512} & \textbf{1024} & \textbf{2048} & \textbf{4096} & \textbf{512} & \textbf{1024} & \textbf{2048} & \textbf{4096} & \textbf{Dense} & $\bm{\Delta}_{\textbf{2048}}$ \\
\midrule
StarCoder & 62.34 & 76.28 & 84.12 & 90.73 & 0.5789 & 0.5717 & 0.5699 & 0.5703 & 0.5703 & $-$0.0004 \\
Books     & 51.37 & 65.11 & 76.29 & 84.10 & 1.7720 & 1.7560 & 1.7500 & 1.7480 & 1.7446 & +0.0054 \\
ArXiv     & 55.77 & 67.73 & 80.67 & 89.55 & 1.1070 & 1.0900 & 1.0844 & 1.0837 & 1.0818 & +0.0026 \\
\bottomrule
\end{tabular}
\end{table}

\Cref{tab:attention_analysis} summarizes attention behavior on books, ArXiv, and StarCoder data. As we increase the number of activated tokens, sparse attention recovers a growing fraction of the dense attention mass. At a $1{:}16$ activation ratio, which corresponds to $2048$ selected tokens, the routed set already captures roughly $80\%$ of the dense attention score, showing that a small active subset can approximate most of the global allocation.

Second, matching full dense attention is not the same as matching language-modeling quality. Across Books, ArXiv, and StarCoder, sparse attention reaches dense-level quality, with loss differences around $0.006$ or smaller. On StarCoder, the sparse curve even slightly improves over dense YOCO. This shows that selected-token sparse attention can preserve dense quality without recovering $100\%$ of the attention mass.

This analysis also clarifies the relationship among block sparse attention, token sparse attention, and shared routing. Block sparse attention gains efficiency from structured memory access, but it relies on a block-level inductive bias that is not well aligned with long-context retrieval. Nearby tokens may serve very different semantic roles and exhibit very different attention patterns, so a block can contain both irrelevant tokens and a few crucial ones. This makes it difficult for block sparsity to reach the same long-context quality as fine-grained selection. Token sparse attention avoids this bias by estimating saliency at token granularity, but recomputing token-level top-$k$ independently in every layer is expensive. \oursattn{} keeps the token-level routing index and shares it across layers. The validity of this sharing comes from the empirical similarity of cross-layer attention scores induced by the data distribution and shared KV memory, rather than from an imposed block structure, which helps preserve quality while amortizing routing cost.

\section{Related Work}

\subsection{Sparse Attention}
Training-aware approaches including NSA~\cite{nsa} and MoBA~\cite{moba} integrate dynamic sparse attention directly into model training. A parallel line of post-training or training-free methods~\cite{seerattention-r,resa,seerattention,quest,deepseek3.2} exploits similar sparsity patterns at inference time. These works generally reduce the quadratic cost of dense attention by adaptively selecting a subset of relevant tokens or blocks, thereby improving efficiency while retaining strong model quality. In contrast to these methods, a central challenge in prior dynamic sparse attention methods is the efficiency-quality trade-off. Methods with stronger accuracy typically yield limited end-to-end speedup, while methods that achieve more aggressive acceleration often incur noticeable quality degradation.

A second line of work exploits the observation that salient tokens tend to remain relatively stable across nearby transformer layers. Some methods use this property in a training-free manner, relying on heuristic cross-layer propagation of token importance patterns during inference~\cite{kascade,omnikv,tidaldecode,delta}. More recently, training-aware approaches such as HySparse~\cite{hysparse} and IndexCache~\cite{indexcache} explicitly learn cross-layer salient-token reuse, through post-training distillation or end-to-end training. These works provide evidence that cross-layer saliency reuse is often well aligned with the ground-truth attention patterns. However, this observation is largely established in conventional attention architectures, where the resulting speedup is often limited in practice. In contrast, we incorporate this idea into YOCO architecture for improving decode efficiency, while the pre-filling and KV cache storage remain efficient.

\subsection{Hybrid Architecture}
A closely related line of work studies hybrid architectures that combine attention with more efficient sequence operators, such as Mamba~\cite{mamba}, RetNet~\cite{retnet}, and Gated DeltaNet~\cite{gdn}. In practice, hybrid models integrate these operators with softmax attention using carefully designed schedules~\cite{jamba,minimax01,qwen3,kimi-linear}. Overall, these works suggest that hybrid scheduling can improve efficiency without substantially harming model quality.

A related line of work combines hybrid architectures with cross-layer KV cache sharing to further reduce memory cost. YOCO~\cite{yoco} introduces this design into the architectural level, allowing later layers to reuse KV states from earlier layers rather than maintaining fully independent caches. Subsequent works adopt closely related ideas in a range of model families, including CLA~\cite{cla}, Gemma 3n~\cite{gemma3}, and Phi-4-mini-Flash~\cite{phi4miniflash}. Collectively, these studies suggest that cross-layer KV reuse can substantially reduce cache memory with limited impact on model quality. However, these works primarily target prefill efficiency and KV cache memory reduction through cross-layer KV reuse, whereas our focus is on improving decoding efficiency in which dense layers identify informative context and subsequent sparse layers reuse that information in a structured manner.

\section{Conclusion}

We presented cross-layer sparse attention for KV-sharing architectures that extends sharing from memory to routing. By reusing one shared routing index across cross-decoder layers, \oursattn{} preserves the fine-grained selectivity of token sparse attention while amortizing the practical cost of top-$k$ routing. This yields an architecture that jointly improves the three major inference bottlenecks in long-context LLMs, namely pre-filling, KV-cache storage, and decoding. Empirically, \oursattn{} remains nearly lossless across short-context and long-context evaluations while delivering substantial end-to-end acceleration, including up to $7.6\times$ decode speedup and $17.1\times$ overall throughput improvement at 128K context. We hope this result provides a more complete architectural direction for long-context LLMs that better reconciles model quality and inference efficiency.

\bibliographystyle{abbrvnat}
\bibliography{clsa}

\newpage
\appendix

\section{Dense-Stage Training Curves}
\label{app:dense-training-curves}
\begin{figure}[h]
\centering
\includegraphics[width=\linewidth]{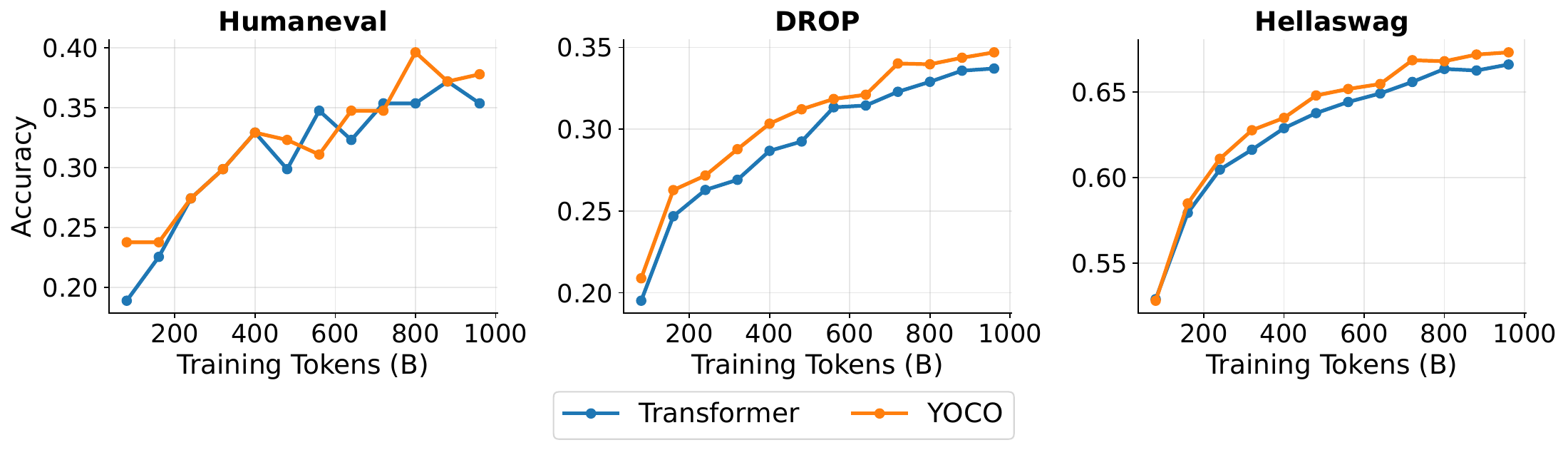}
\caption{Dense-stage training curves on HumanEval, DROP, and HellaSwag as a function of training tokens. YOCO remains competitive with the Transformer throughout training, supporting its use as a stable dense backbone before sparse adaptation.}
\label{fig:updates}
\end{figure}

\Cref{fig:updates} shows the dense-stage training curves on representative benchmarks. YOCO remains competitive with the Transformer throughout training. Since \oursattn{} is coupled with the YOCO backbone, these curves also serve as a sanity check that YOCO provides a strong dense attention starting point, including on retrieval-style tasks such as DROP. This makes it possible to start from a good dense model and obtain our final model through a near-lossless sparse-attention adaptation, rather than relying on a substantially different pretraining recipe.

\section{Training Hyper-parameters}
\label{app:training-hparams}
This appendix provides the exact optimization settings used for dense pretraining and sparse adaptation. Dense pretraining follows a two-stage schedule: stage 1 trains with a peak learning rate of $3{\times}10^{-4}$ using 2000 warmup iterations on 8K contexts, and stage 2 increases the context cap to 32,768 while switching to a fixed learning rate of $3{\times}10^{-5}$ for the remaining 10,000 updates. Sparse adaptation then reuses the same training recipe on 32,768-token sequences: stage 1 keeps the batch size and peak learning rate from dense stage 1, and uses the warmup iterations listed below; stage 2 continues for another 2,500 updates at the smaller fixed learning rate.
For completeness, we also include the shared optimizer settings (Adam betas, epsilon, and weight decay), which are applied across all stages unless explicitly overridden.

\begin{table}[h]
\centering
\small
\setlength{\tabcolsep}{4pt}
\caption{Stage-wise training hyperparameters for dense pretraining and sparse adaptation.}
\label{tab:train_hparams}
\vspace{0.5em}
\begin{tabular}{lcccc}
\toprule
& \textbf{Dense Stage 1} & \textbf{Dense Stage 2} & \textbf{Sparse Stage 1} & \textbf{Sparse Stage 2} \\
\midrule
Learning rate & $3{\times}10^{-4}$ & $3{\times}10^{-5}$ & $3{\times}10^{-4}$ & $3{\times}10^{-5}$ \\
Minimum LR & $3{\times}10^{-5}$ & $3{\times}10^{-5}$ & $3{\times}10^{-4}$ & $3{\times}10^{-5}$ \\
Max sequence length & $8192$ & $32768$ & $32768$ & $32768$ \\
Warmup iterations & $2000$ & 0 & 500 & 0 \\
Training steps & $125000$ & $10000$ & $2500$ & $2500$ \\
\bottomrule
\end{tabular}
\end{table}

\begin{table}[h]
\centering
\small
\caption{Shared optimization settings across all training stages.}
\label{tab:train_hparams_shared}
\vspace{0.5em}
\begin{tabular}{lc}
\toprule
\textbf{Hyper-parameter} & \textbf{Value} \\
\midrule
Batch size & 8M \\
Adam $\beta$ & ($0.9$, $0.95$) \\
Adam $\varepsilon$ & $10^{-8}$ \\
Weight decay & $0.1$ \\
\bottomrule
\end{tabular}
\end{table}

\section{Model Configuration}
\label{app:model-config}

Tables~\ref{tab:model_config_shared} and~\ref{tab:model_config_diff} list the architectural hyperparameters for all evaluated variants. Across models we keep the same overall width and depth, and (for the YOCO-based models) split the 32 layers into 16 self-decoder layers and 16 cross-decoder layers. The key practical difference lies in how positional information is handled: the Transformer applies RoPE throughout, while YOCO (\oursattn{}) uses an RNoPE setting that restricts RoPE to the sliding-window self-decoder and removes positional encoding from the global cross-decoder attention path.
We additionally enable QK normalization and use GQA in all models, as summarized in the table.

\begin{table}[h]
\centering
\small
\caption{Shared architectural hyperparameters across all evaluated models.}
\label{tab:model_config_shared}
\vspace{0.5em}
\begin{tabular}{lc}
\toprule
\textbf{Hyperparameter} & \textbf{Value} \\
\midrule
Hidden size & $2560$ \\
FFN width & $7680$ \\
Layers & $32$ \\
Heads & $20$ \\
KV heads & $4$ \\
Head dimension & $128$ \\
QK Norm & enabled \\
Weight tying & disabled \\
\bottomrule
\end{tabular}
\end{table}

\begin{table}[h]
\centering
\small
\setlength{\tabcolsep}{3pt}
\caption{Key architectural differences among the Transformer baseline, YOCO (Dense), and YOCO (\oursattn{}), including positional encoding and attention type.}
\label{tab:model_config_diff}
\vspace{0.5em}
\begin{tabular}{lccc}
\toprule
& \textbf{Transformer} & \textbf{YOCO (Dense)} & \textbf{YOCO (\oursattn{})} \\
\midrule
Positional encoding & RoPE & RNoPE & RNoPE \\
RoPE base & $5\times 10^5$ & $1\times 10^4$ & $1\times 10^4$ \\
Attention type & GQA & GQA & GQA+\oursattn{} \\
\bottomrule
\end{tabular}
\end{table}

\section{Experimental Details of Latency Breakdown}
\label{app:exp-details-breakdown}
This section lists the per-layer latency values used in \Cref{fig:breakdown}. The table reports the plotted values in milliseconds after the script's scaling and per-layer averaging. For YOCO (Dense), the reported attention term averages SWA in the 16 self-decoder layers and dense full attention in the 16 cross-decoder layers. For YOCO (\oursattn{}), it averages the same 16 SWA layers and 16 CLSA layers. The top-$k$ term is the amortized per-layer routing cost, obtained by dividing the one-off routing latency by the full 32-layer model depth to match the normalization used in the plot. The routing decision itself is shared by the 16 cross-decoder layers, and the unamortized one-off 128K top-$k$ cost is shown separately in \Cref{fig:breakdown_128k}.

\begin{table}[h]
\centering
\small
\setlength{\tabcolsep}{5pt}
\caption{Per-layer latency breakdown (ms) used in \Cref{fig:breakdown}. For the YOCO variants, attention terms are averaged across layer types, and the top-$k$ column reports the amortized per-layer routing cost for \ours{}.}
\label{tab:fig_breakdown_data}
\vspace{0.5em}
\begin{tabular}{llccccc}
\toprule
\textbf{Context} & \textbf{Model} & \textbf{MLP} & \textbf{Dense Attn.} & \textbf{Sparse Attn.} & \textbf{Top-$k$} & \textbf{Total} \\
\midrule
8K & Transformer & 0.11 & 0.14 & -- & -- & 0.25 \\
8K & YOCO (Dense) & 0.11 & 0.07 & -- & -- & 0.18 \\
8K & YOCO (\oursattn{}) & 0.11 & -- & 0.03 & 0.01 & 0.15 \\
\midrule
32K & Transformer & 0.13 & 0.50 & -- & -- & 0.63 \\
32K & YOCO (Dense) & 0.13 & 0.24 & -- & -- & 0.37 \\
32K & YOCO (\oursattn{}) & 0.13 & -- & 0.03 & 0.02 & 0.18 \\
\midrule
128K & Transformer & 0.17 & 2.11 & -- & -- & 2.28 \\
128K & YOCO (Dense) & 0.17 & 0.87 & -- & -- & 1.04 \\
128K & YOCO (\oursattn{}) & 0.17 & -- & 0.05 & 0.08 & 0.31 \\
\bottomrule
\\
\end{tabular}
\end{table}

\section{Experimental Details of Inference Throughput}
\label{app:exp-details-inference}

This section lists the raw throughput measurements used to produce the plotted panels of \Cref{fig:inference}, and also reports overall end-to-end generation throughput for the same setup. In the main paper we plot throughput ratios relative to the Transformer; the absolute tokens/s values reported here correspond to the same measurements for prefill, decode, and overall end-to-end generation (all evaluated on NVIDIA B200 GPUs).

\begin{table}[h]
\centering
\small
\setlength{\tabcolsep}{6pt}
\caption{Raw prefill throughput (tokens/s) used in the left panel of \Cref{fig:inference}.}
\label{tab:fig_inference_prefill}
\vspace{0.5em}
\begin{tabular}{lccc}
\toprule
\textbf{Context Length} & \textbf{Transformer} & \textbf{YOCO (Dense)} & \textbf{YOCO (\oursattn{})} \\
\midrule
8K & 4721.71 & 10884.35 & 9623.48 \\
16K & 4160.92 & 13033.56 & 11572.40 \\
32K & 2889.39 & 18450.18 & 18163.24 \\
64K & 1741.56 & 20343.94 & 20349.11 \\
128K & 1019.06 & 20864.85 & 20741.51 \\
\bottomrule
\\
\end{tabular}
\end{table}

\begin{table}[h]
\centering
\small
\setlength{\tabcolsep}{6pt}
\caption{Raw decode throughput (tokens/s) used in the right panel of \Cref{fig:inference}.}
\label{tab:fig_inference_decode}
\vspace{0.5em}
\begin{tabular}{lccc}
\toprule
\textbf{Context Length} & \textbf{Transformer} & \textbf{YOCO (Dense)} & \textbf{YOCO (\oursattn{})} \\
\midrule
8K & 4762.79 & 5516.50 & 6742.39 \\
16K & 3091.32 & 4147.85 & 6350.39 \\
32K & 1761.72 & 2677.12 & 5461.33 \\
64K & 948.15 & 1600.00 & 4137.37 \\
128K & 431.16 & 960.94 & 3276.80 \\
\bottomrule
\\
\end{tabular}
\end{table}

\begin{table}[h]
\centering
\small
\setlength{\tabcolsep}{6pt}
\caption{Raw overall throughput (tokens/s) measured under the same setup as \Cref{fig:inference}.}
\label{tab:fig_inference_overall}
\vspace{0.5em}
\begin{tabular}{lccc}
\toprule
\textbf{Context Length} & \textbf{Transformer} & \textbf{YOCO (Dense)} & \textbf{YOCO (\oursattn{})} \\
\midrule
8K & 2989.78 & 4311.58 & 4920.12 \\
16K & 1449.91 & 2904.96 & 3715.19 \\
32K & 570.27 & 1832.66 & 2805.48 \\
64K & 191.36 & 1042.90 & 1735.59 \\
128K & 62.53 & 599.05 & 1068.06 \\
\bottomrule
\\
\end{tabular}
\end{table}

% \newpage
% \input{checklist.tex}

\end{document}